\pdfoutput=1

\documentclass[11pt]{article}

\usepackage{acl}

\usepackage{times}
\usepackage{latexsym}
\usepackage{multirow}
\usepackage{graphicx}
\usepackage{pdfpages}
\usepackage[T1]{fontenc}

\usepackage[utf8]{inputenc}

\usepackage{microtype}

\usepackage{booktabs}
\usepackage{xcolor}
\usepackage{pifont}
\usepackage{amsmath,amssymb,amsfonts}%
\usepackage{amsthm}%
\usepackage{mathrsfs}%
\usepackage{textcomp}%
\usepackage{manyfoot}%
\usepackage{listings}%
\usepackage{subcaption}
\usepackage{caption}
\usepackage{pgf-pie}  
\usepackage{url}

\definecolor{myRed}{rgb}{0.808,0.067,0.149}
\definecolor{myGreen}{rgb}{0.067,0.708,0.149}

\title{RoDia: A New Dataset for Romanian Dialect Identification from Speech}

\author{Codru\c{t} Rotaru$^1$ \and Nicolae C\u{a}t\u{a}lin Ristea$^{1,2}$ \and Radu Tudor Ionescu$^1$\\
$^1$University of Bucharest, Romania\\
$^2$National University of Science and Technology Politehnica Bucharest, Romania\\
Corresponding author: \texttt{raducu.ionescu@gmail.com}}
\begin{document}
\maketitle
\begin{abstract}
We introduce RoDia, the first dataset for Romanian dialect identification from speech. The RoDia dataset includes a varied compilation of speech samples from five distinct regions of Romania, covering both urban and rural environments, totaling 2 hours of manually annotated speech data. Along with our dataset, we introduce a set of competitive models to be used as baselines for future research. The top scoring model achieves a macro $F_1$ score of $59.83\%$ and a micro $F_1$ score of $62.08\%$, indicating that the task is challenging. We thus believe that RoDia is a valuable resource that will stimulate research aiming to address the challenges of Romanian dialect identification. We release our dataset at \url{https://github.com/codrut2/RoDia}.
\end{abstract}


\section{Introduction}

Spoken dialect identification emerged as a challenging task aiming to achieve a fine-grained distinction between varieties of a certain language, having similar implications to spoken language identification \cite{Barnard-SLTU-2014,Kimanuka-ICLR-2023,Ma-TASLP-2007}. Despite being a more delicate task, spoken dialect identification received comparatively lower attention, most of it being devoted to dialect identification for widely spoken languages, such as English \cite{Weinberger-Brill-2011}, Chinese \cite{Zhang-ISCSLP-2022}, and Arabic \cite{Ali-ASRU-2017,Ali-ASRU-2019,Shon-ICASSP-2020}. Spoken dialect identification for low-resource languages, such as Swiss German \cite{Dogan-ArXiv-2021,Pluss-ACL-2023} and Finnish \cite{Hamalainen-EMNLP-2021}, has remained relatively underexplored \cite{Ranathunga-CSUR-2023, Barnard-SLTU-2014, Hamalainen-EMNLP-2021}. Different from prior studies, we focus on spoken language identification in Romanian, a low-resource language characterized by its intricate dialectal variations within the country of Romania.
Romanian, a Romance language with Latin roots, boasts a rich linguistic landscape shaped by historical, geographical, and sociocultural factors \cite{mititelu2018reference}. 
However, despite its linguistic complexity, Romanian remains a low-resource language, with limited studies dedicated to understanding its regional linguistic diversity. This scarcity of resources is not unique to Romanian, numerous other languages around the world having similar challenges due to their lower visibility on the global linguistic stage \cite{Ranathunga-CSUR-2023, Barnard-SLTU-2014, Hamalainen-EMNLP-2021}. Notably, the VarDial workshop is one of the main drivers for growing the interest around language variety and dialect identification, through the organization of multiple shared tasks each year \cite{Aepli-VarDial-2022,Chakravarthi-VarDial-2021,Gaman-VarDial-2020,Zampieri-VarDial-2019}.

Due to the success of deep learning frameworks in speech processing \cite{mehrish2023review}, researchers started to employ such methods in the area of low-resource languages \cite{chan2015deep, al2023automatic}. This has led to a growing need for resources on low-resource languages. Considering dialect identification datasets across different languages, we can distinguish between two types of resources: text-based datasets \cite{Bouamor-LREC-2018,Butnaru-ACL-2019,Francom-LREC-2014,Gaman-KES-2023,Gaman-IJIS-2022} and speech-based datasets \cite{Ali-ASRU-2017,Ali-ASRU-2019,Shon-ICASSP-2020,Dogan-ArXiv-2021,Pluss-ACL-2023, Hamalainen-EMNLP-2021}. While various languages have benefited from text-based resources that leverage written materials capturing linguistic variations, the auditory dimension of dialects adds an intricate layer of complexity. Text data, although valuable, might not fully encapsulate the nuanced phonetic and prosodic characteristics that are pivotal in dialect differentiation. 
In contrast to text datasets, audio datasets \cite{Ali-ASRU-2017,Ali-ASRU-2019,Shon-ICASSP-2020,Dogan-ArXiv-2021,Pluss-ACL-2023} offer a more holistic representation, capturing not only the lexical disparities, but also the subtle intonations and accents inherent in speech.

\begin{figure}[!t]
\begin{center}
\centerline{\includegraphics[width=1.0\linewidth]{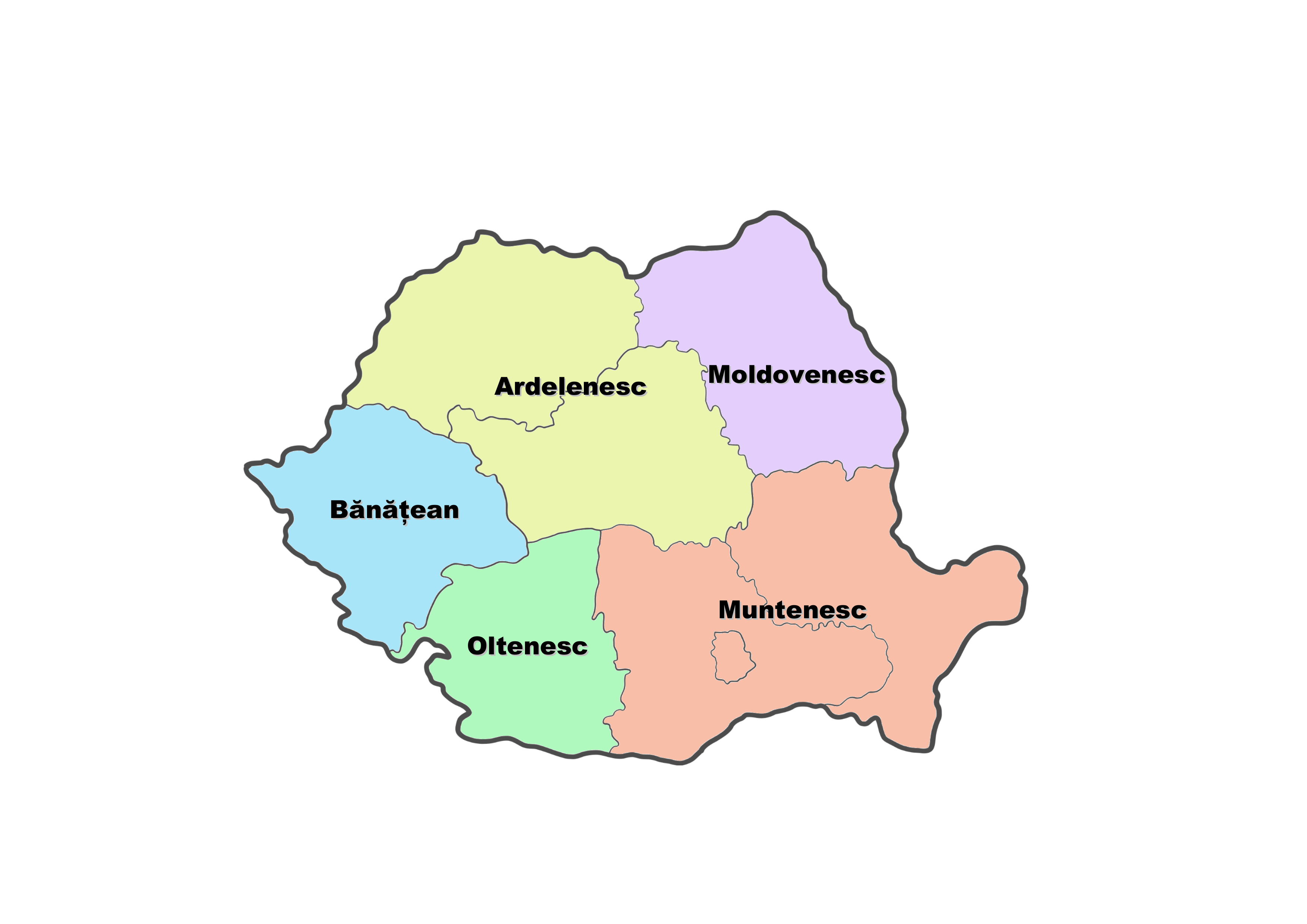}}
\vspace{-0.2cm}
\caption{The administrative regions of Romania and the dominant dialect spoken within each region. RoDia is the first benchmark to contain samples representing these five Romanian dialects.}
\label{fig_Romania}
\vspace{-0.8cm}
\end{center}
\end{figure}

To the best of our knowledge, RoDia is the first dataset to tackle spoken dialect identification in the Romanian landscape in accordance with historical, geographical, and sociocultural factors, encouraging the research in this low-resource language. Although there are two text datasets addressing Romanian dialect identification, MOROCO \cite{Butnaru-ACL-2019} and MOROCO-Tweets \cite{Gaman-IJIS-2022}, these cover only two dialects: Romanian (equivalent to the \emph{Muntenesc} dialect) and Moldavian (\emph{Moldovenesc}). In contrast, our dataset is focused on speech and covers five Romanian dialects, as shown in Figure \ref{fig_Romania}. We underline that the extra dialects, namely \emph{Ardelenesc}, \emph{B\u{a}n\u{a}\c{t}ean}, and \emph{Oltenesc}, are very well characterized by phonetic differences captured only in speech. This explains why MOROCO \cite{Butnaru-ACL-2019} and MOROCO-Tweets \cite{Gaman-IJIS-2022} only contain text samples from the other two dialects.

The number of audio datasets available in Romanian is rather low \cite{Avram-COMM-2022,Georgescu-LREC-2020}, confirming that Romanian is indeed a low-resource language. We underline that existing datasets comprising Romanian speech samples are mainly focused on automatic speech recognition, ignoring the diversity of dialects within the region. In contrast, our dataset is specifically designed to represent five distinct dialects spoken in Romania: \textit{Muntenesc} (accepted as the official language of Romania), \textit{Ardelenesc}, \textit{Moldovenesc}, \textit{Oltenesc}, and \textit{B\u{a}n\u{a}\c{t}ean}\footnote{We refer to the original (untranslated) dialect names, since most of them have no translation in English.}. 

\section{Dataset}


\noindent
\textbf{Data collection and annotation.}
We collected the vast majority of the audio samples by gathering interviews and shows from local TV channels from all five regions considered in the dataset (see Figure \ref{fig_Romania}). To obtain a clean dataset, we employed a rigorous selection process. First, we manually cropped the gathered audio files with respect to each speaker, e.g.~we split an interview into multiple samples, such that each sample contains a single speaker. Next, we discard samples with interfering speakers and with a low perceived intelligibility. To make sure the label assignment is robust, we submitted all samples gathered from the TV channels to local annotators to validate the assigned labels. The manual validation process eliminates samples with a questionable dialect. In addition, a small proportion of data was acquired by recording citizens native to the five regions, who were asked to read some random texts from the Romanian Wikipedia, in their own dialect. We cropped the recorded samples to minimize the amount of silence at the start and the end of each audio sample. Upon curating the gathered data, we are left with a clean dataset containing $2,\!768$ audio samples, each having between $2.5$ and $5.0$ seconds of speech. The sample rate of all samples is $44.1$ kHz.

We divide the dataset into $2,\!164$ samples for training and $604$ samples for testing, such that there is no overlap between speakers in training and test. Without separating the speakers between training and test, a model that overfits to certain speaker-specific features that are not related to dialect (e.g.~pitch, loudness, rate, etc.) will reach high scores on the test set. However, these scores are unlikely to represent the actual performance of the model in a real-world scenario, where the audio samples come from unknown speakers. We thus consider that a more realistic evaluation is to separate the speakers. In the proposed setting, models that learn patterns related to speakers will not be able to capitalize on features unrelated to dialect identification.

\begin{table*}[!t]
\small{
  \begin{center}
  \begin{tabular}{|l|c|c|c|c|c|c|c|}
    \hline
    \multirow{2}{*}{Class} & \multirow{2}{*}{\#speakers} & \multicolumn{3}{c|}{Train} & \multicolumn{3}{c|}{Test} \\
    \cline{3-8}
    & & \#samples  & SNR & SRR & \#samples  & SNR & SRR  \\
    \hline
    \hline
    Ardelenesc & 47 & 427 & 28.8 & 36.4 & 119 & 30.5 & 38.5\\
    B\u{a}n\u{a}\c{t}ean & 67 & 424 & 23.1 & 34.6 & 99 & 25.0 & 37.7\\
    Moldovenesc & 47 & 384 & 25.6 & 32.4 & 206 & 25.7 & 27.8\\
    Muntenesc & 64 & 603 & 29.0 & 35.3 & 106 & 26.7 & 37.8\\
    Oltenesc & 31 & 326 & 26.6 & 31.2 & 74 & 26.5 & 33.7 \\
    \hline
    Overall & 256 & 2164 & 26.9 & 34.2 & 604 & 26.8 & 34.0\\
    \hline
  \end{tabular}
  \end{center}
    \vspace{-0.2cm}
  \caption{The number of training and test samples for each class in our dataset. For reference, we include the average SNR and SRR values for each category. The number of speakers per dialect is also provided.}
\label{dataset_tab}
}
\vspace{-0.1cm}
\end{table*}

There are five local annotators (one per region), who annotated all samples. The inter-rater Quadratic Weighted Kappa score is 0.83, indicating that the collected labels exhibit a substantial agreement among human evaluators. The average accuracy of our raters is 86\%, indicating that the task is fairly easy for humans. Note that all raters are native Romanian speakers who speak the literary language, as well as at least one of the five dialects.

The collected samples comprise interviews and read speech found on YouTube. The read speech is actually gathered from videos where various speakers read from different books (without any influence or preparation from our side). The percentage of read speech is 21\%.

Aside from dialect labels, our annotators also labeled each audio sample with the gender and age of each speaker. More precisely, the age and gender of each speaker is estimated by two annotators who had to analyze both video and audio modalities. The age annotation consists in classifying each speaker into a 10-year age group, after watching the video available for the respective speaker. In summary, our audio samples come with dialect, age and gender labels, enabling the study of additional tasks such as gender prediction or age estimation from speech.

\noindent
\textbf{Dataset statistics.}
For a more comprehensive view, we present both demographic information and audio quality statistics for our new dataset. In Table \ref{dataset_tab}, we report the number of samples for each dialect in both train and test splits, as well as the signal-to-noise ratio (SNR) and signal-to-reverberation ratio (SRR). 
Regarding data quality, we note that the SNR and SRR values are consistently higher than $23$ dB, highlighting that the audio samples have relatively low noise and reverberation. The \textit{Muntenesc} dialect has the largest number of audio samples. This dialect was easy to collect, since it represents the literary language, which is often borrowed by speakers native to other regions. On the opposite side, the \textit{Oltenesc} dialect is least represented, having only 400 audio samples. However, the distribution gap between the five classes is not high enough to pose significant challenges to machine learning models. 

\begin{figure}
    \begin{tikzpicture}
    \tikzstyle{every node}=[font=\small]
    \pie[scale=0.28, yshift=1.2cm]{40.5/Female, 59.5/Male}
    \hspace{0.3cm}
    \pie[scale=0.53, xshift=6.7cm]{5.0/{10-20}, 17.9/20-30, 12.6/30-40, 19.0/40-50, 24.6/50-60, 20.9/60+}
\end{tikzpicture}
\vspace{-0.25cm}
\caption{Age and gender statistics for the RoDia dataset.}
\vspace{-0.3cm}
\label{fig_demographics}
\end{figure}
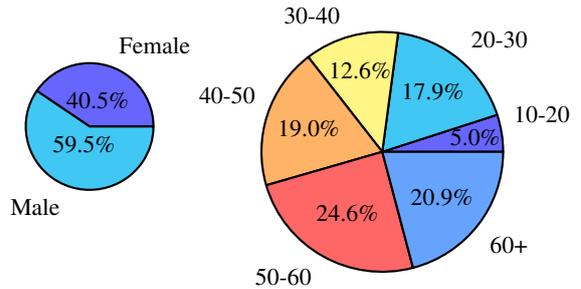

We present demographic information in Figure \ref{fig_demographics}. In terms of demographic insights, RoDia exhibits a relatively balanced gender distribution, having $59.5\%$ male and $40.5\%$ female speakers. Aside from separating the speakers between training and test, we also made sure to have similar demographics for the train and test splits, reducing unnecessary distribution gaps. In summary, we consider that RoDia is a suitable resource for spoken dialect identification.


\begin{table*}[!t]
\small{
  \begin{center}
  \setlength\tabcolsep{1.5pt}
  \begin{tabular}{|l|c|c|c|c|c|c|c|c|c|c|c|c|c|c|c|c|c|}
    \hline
    
    \multirow{2}{*}{Model} & \multicolumn{3}{c|}{Ardelenesc} & \multicolumn{3}{c|}{B\u{a}n\u{a}\c{t}ean} & \multicolumn{3}{c|}{Moldovenesc} & \multicolumn{3}{c|}{Muntenesc} & \multicolumn{3}{c|}{Oltenesc} & \multicolumn{2}{c|}{Overall $F_1$} \\
    \cline{2-18}
    & $P$  & $R$ & $F_1$ & $P$  & $R$ & $F_1$ & $P$  & $R$ & $F_1$ & $P$  & $R$ & $F_1$ & $P$  & $R$ & $F_1$  & Micro & Macro \\
    \hline
    \hline
    ResNet-18 & 59.45 & \color{blue}{73.94} & 65.91   & \color{blue}{47.54} & 58.58 & \color{blue}{52.48}     & 73.98 & \color{blue}{62.13} & 67.54     & 44.11 & 56.60 & 49.58     & 88.00 & 29.72 & 44.44               & 58.94 & 55.99 \\
    
    AST & 66.92 & 73.10 & 69.87    & 43.62 & 65.65 & 52.41     & 77.92 & 58.25 & 66.66     & 48.96 & 66.98 & 56.57     & \color{blue}{92.30} & \color{blue}{32.43} & \color{blue}{48.00}                 & 60.76 & 58.70 \\
    
    SepTr & 67.71 & 72.26 & \color{blue}{69.91}    & 41.31 & \color{blue}{69.69} & 51.87     & \color{blue}{81.75} & 58.73 & \color{blue}{68.36}     & 53.33 & 67.92 & 59.75     & 85.18 & 31.08 & 45.54         & 61.42 & 59.08\\
    
    wav2vec 2.0 & \color{blue}{68.00} & 71.42 & 69.67  & 43.13 & 66.66 & 52.38     & 77.63 & 60.67 & 68.10      & \color{blue}{54.34} & \color{blue}{70.75} & \color{blue}{61.47}     & 88.88 & 32.24 & 47.52        & \color{blue}{62.08} & \color{blue}{59.83} \\    
    \hline

  \end{tabular}
  \end{center}
  \vspace{-0.2cm}
   \caption{Spoken dialect identification results of ResNet-18 \cite{He-CVPR-2016}, AST \cite{Gong-INTERSPEECH-2021}, SepTr \cite{Ristea-INTERSPEECH-2022} and wav2vec 2.0 \cite{Baevski-NeurIPS-2020} on the RoDia test set. For a comprehensive evaluation, we report both per dialect and overall results. The best score on each column is highlighted in {\color{blue} blue}.}
\label{results_tab}
}
\vspace{-0.1cm}
\end{table*}

\section{Experiments}


\noindent
\textbf{Baseline methods.}
We compile a lineup of four state-of-the-art neural architectures for speech processing to form a set of competitive dialect identification baselines for our novel dataset. We consider both convolutional \cite{He-CVPR-2016} and transformer-based neural networks \cite{Gong-INTERSPEECH-2021, Ristea-INTERSPEECH-2022}, as well as a hybrid architecture \cite{Baevski-NeurIPS-2020}. We employ the ResNet-18 \cite{He-CVPR-2016} convolutional network, as it was previously used for audio classification tasks \cite{Ristea-IS-2020}. Additionally, we explore two transformer-based architectures, namely the Audio Spectrogram Transformer (AST) \cite{Gong-INTERSPEECH-2021} and the Separable Transformer (SepTr) \cite{Ristea-INTERSPEECH-2022}, due to their high performance in audio classification. We also employ the wav2vec 2.0 model \cite{Baevski-NeurIPS-2020}, which uses a hybrid architecture combining the advantages of both convolutional and transformer blocks. 
All models are trained in the multi-class setting, since the ground-truth labels are constructed in a similar manner: one audio sample belongs to only one dialect.

\noindent
\textbf{Preprocessing.}
For models operating in the time-frequency domain \cite{He-CVPR-2016, Gong-INTERSPEECH-2021, Ristea-INTERSPEECH-2022}, we apply the Short-Time Fourier Transform with a window size of 512 and a hop size of 256. Then, we compute the square root of the magnitude, obtaining the spectrogram map. The other steps and parameters are exactly as described in the original papers. For wav2vec 2.0 \cite{Baevski-NeurIPS-2020}, we apply the preprocessing steps described by the authors, which are mainly used for normalization. 
In all our experiments, we use the following data augmentation methods: noise perturbation, time shifting, speed perturbation, mix-up and SpecAugment \cite{Park-INTERSPEECH-2019}.

\noindent
\textbf{Evaluation metrics.}
We report the precision ($P$), recall ($R$), and $F_1$ scores computed for each dialect. These metrics provide insights into the ability of models to correctly classify instances within each class. To quantify the overall performance, we aggregate the individual scores via the micro and macro $F_1$ measures. The micro $F_1$ score combines the performance metrics across all examples, while the macro $F_1$ score offers a balanced average of the $F_1$ scores across all classes. 

\noindent
\textbf{Training environment.}
All models are optimized on an Nvidia GeForce GTX 3090 GPU with 24 GB of VRAM.

\noindent
\textbf{Hyperparameter tuning.}
For each model, we employed grid search to find the optimal learning rate (between $10^{-2}$ and $10^{-6}$) and the optimal batch size (between $8$ and $128$ samples). We take the wav2vec 2.0 \cite{Baevski-NeurIPS-2020} model, which is pretrained on English data, and fine-tune it for $10$ epochs on RoDia using a learning rate of $10^{-5}$ and mini-batches of $16$ samples. We train ResNet-18, AST \cite{Gong-INTERSPEECH-2021} and SepTr \cite{Ristea-INTERSPEECH-2022} from scratch. The models are trained for $50$ epochs with early stopping, using a learning rate of $10^{-4}$ and mini-batches of $32$ samples. All models are optimized with the Adam optimizer \cite{Kingma-ICLR-1015}. 


\begin{figure}[!t]
\begin{center}
\centerline{\includegraphics[width=0.9\linewidth]{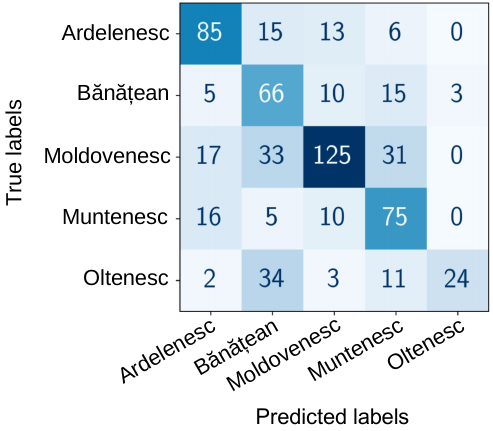}}
\vspace{-0.2cm}
\caption{Confusion matrix on the test set for the wav2vec 2.0 \cite{Baevski-NeurIPS-2020} model.}
\label{fig_confusion}
\vspace{-0.8cm}
\end{center}
\end{figure}

\noindent
\textbf{Dialect identification results.}
In Table \ref{results_tab}, we present the spoken dialect identification results of the baseline models on the RoDia test set. The convolutional network, ResNet-18, obtains the lowest overall performance. Still, ResNet-18 reaches competitive $F_1$ scores for the \emph{B\u{a}n\u{a}\c{t}ean} and \emph{Moldovenesc} dialects. Unlike the other baselines, the ResNet-18 model struggles with the \emph{Muntenesc} and \emph{Ardelenesc} dialects, which explains its low overall performance. The transformer-based models, AST \cite{Gong-INTERSPEECH-2021} and SepTr \cite{Ristea-INTERSPEECH-2022}, yield superior results, with a slight upper hand from the SepTr model. In terms of the overall $F_1$ scores, the best model appears to be wav2vec 2.0 \cite{Baevski-NeurIPS-2020}. However, the $F_1$ scores per dialect seem to tell a slightly different story, since the wav2vec 2.0 is outperformed by at least one of the other models on four dialects: \emph{Ardelenesc}, \emph{B\u{a}n\u{a}\c{t}ean}, \emph{Moldovenesc} and \emph{Oltenesc}. The competitive edge of wav2vec 2.0 lies in its ability to better identify the \emph{Muntenesc} dialect. We underline that the audio samples were recorded in uncontrolled scenarios, so the reported results directly reflect the capability of systems in the real-world case.

To further assess the behavior of the best baseline model, namely wav2vec 2.0, we consider its confusion matrix illustrated in Figure \ref{fig_confusion}. The confusion matrix reveals some interesting patterns. While the model tends to mislabel samples from the \emph{Oltenesc} dialect, we observe that most of these mistakes are caused by a high confusion with the \emph{B\u{a}n\u{a}\c{t}ean} dialect. Since Banat and Oltenia are neighboring regions, there are several similarities between these two dialects. For instance, both dialects are characterized by the frequent use of the perfect simple tense, which is hardly encountered in the other Romanian dialects. Another noticeable problem is with the \emph{Moldovenesc} dialect, which is often wrongly identified as \emph{B\u{a}n\u{a}\c{t}ean} and \emph{Muntenesc}. The confusion between the \emph{Moldovenesc} and \emph{B\u{a}n\u{a}\c{t}ean} dialects is caused by the fact these two dialects kept the form of words such as \emph{c\^{a}ne} (dog), \emph{p\^{a}ne} (bread) and \emph{m\^{a}ne} (tomorrow), from the old Romanian. In the literary language, as well as the other dialects, these words are pronounced with an `i' before the consonant `n', as follows: \emph{c\^{a}ine}, \emph{p\^{a}ine} and \emph{m\^{a}ine}. The confusion with the \emph{Muntenesc} dialect can be attributed to the fact that some residents of the southern part of Moldavia lost some of the dialectal features, e.g.~they prefer to use the word \emph{pantofi} to refer to shoes, and the word \emph{papuci} to refer to slippers, just as the residents of Muntenia. In contrast, the residents of the northern side of Moldavia regularly use \emph{papuci} to refer to shoes, and \emph{\c{s}lapi} to refer to slippers. In summary, the confusion matrix shows that Romanian dialect identification is not an easy task, requiring researchers to address specific issues in order to come up with more accurate models in the future.

The confusion matrix illustrated in Figure \ref{fig_confusion} also shows that the training data distribution does not affect wav2vec 2.0. For example, the Moldovenesc dialect is the second-least popular dialect in our dataset, but wav2vec places many of the test samples into the Moldovenesc class. Overall, the confusion matrix of wav2vec reflects the test data distribution, although the model was trained on a slightly different class distribution. This confirms that the imbalance is not high enough to bias models. 

\begin{table}[!t]
\small{
  \begin{center}
  \begin{tabular}{|l|c|}
    \hline
    {Dialect} & WER \\
    \hline
    \hline
    Ardelenesc & 31.5\%\\
    B\u{a}n\u{a}\c{t}ean & 30.2\%\\
    Moldovenesc & 32.8\%\\
    Muntenesc & 24.1\%\\
    Oltenesc & 29.7\% \\
    \hline
  \end{tabular}
  \end{center}
    \vspace{-0.2cm}
  \caption{Word error rates (WER) of the Whisper-Large model \cite{Radford-ICML-2023} on the five dialects from the RoDia dataset. The Whisper-Large model is trained on the literary Romanian language. The ASR transcripts are compared with manual transcripts to establish the performance of the Whisper-Large model.}
\label{tab_asr}
}
\vspace{-0.1cm}
\end{table}

\noindent
\textbf{Speech recognition results.}
We manually transcribed our data samples to test the performance of a state-of-the-art automatic speech recognition (ASR) system on RoDia. Then, we applied the open source Whisper-Large model \cite{Radford-ICML-2023} on our test set and obtained the word error rates reported in Table \ref{tab_asr}. The \emph{Muntenesc} dialect is almost identical to the literary language, explaining why it exhibits the lowest WER. The WER obtained by the Whisper-Large model for the Romanian language on the Common Voice dataset is 19.8\%. The difference between the WER for the Muntenesc dialect and the WER reported on Common Voice can be attributed to the distribution gap between the RoDia and the Common Voice datasets. Considering the generally higher error rates for the other Romanian dialects shown in Table \ref{tab_asr}, we conclude that ASR for dialectal speech is more difficult. This justifies the utility of our novel dataset for ASR.

\section{Conclusion}

In this paper, we introduced RoDia, the first dataset for Romanian dialect identification from speech. Our dataset contains $2,\!768$ speech samples representing five Romanian dialects. The audio samples were manually annotated with dialect, age and gender labels, enabling the study of spoken dialect identification in a realistic scenario, where the speakers in the training and test splits are disjoint. We conducted experiments with four state-of-the-art speech processing models, establishing a range of baseline performance levels for future research.

\section{Limitations}

This work is focused on spoken Romanian dialect identification, but the performance levels of the considered approaches might be different on other languages. Due to our specific focus on the Romanian language, we did not evaluate the performance of the considered models across other languages. However, we consider that the evaluation on other languages is beyond the scope of the current study.

Another limitation of our work is the slightly limited number of samples with manual labels included in our corpus. This limitation is caused by scarcity of resources available online. Most of the Romanian video or audio samples available online use the literary language. Local news and content creators commonly use the literary language taught in schools. Hence, dialects are mostly used by rural residents, which often have no Internet access. This situation significantly limits the dialectal resources that are publicly available.

\section{Ethics Statement}

The manual labeling was carried out by volunteers who agreed to annotate the audio samples for free. Prior to the annotation, they also agreed to let us publish their labels along with the dataset. Our data is collected from YouTube, which resides in the public web domain. We note that the European regulations\footnote{\url{https://eur-lex.europa.eu/eli/dir/2019/790/oj}} allow researchers to use data in the public web domain for non-commercial research purposes. Thus, we release our data and code under the CC BY-NC-SA 4.0 license\footnote{\url{https://creativecommons.org/licenses/by-nc-sa/4.0/}}.

During data collection, we made sure the audio samples do not contain information that names or uniquely identifies individual people.

\bibliography{anthology,custom}

\clearpage
\appendix

\section{Appendix}
\label{sec:appendix}

\noindent
\textbf{Peculiarities of Romanian dialects.}
As per Wikipedia\footnote{\url{https://en.wikipedia.org/wiki/Romanian_dialects}}, the Romanian dialects are not easy to classify, and their classification is still highly debated by experts, who proposed various classifications, ranging from 2 to even 20 dialects. Since there is no standard classification, we underline that the dialects from RoDia are not unanimously considered as dialects by experts. In Romanian, these are called ``grai'', which is translated (perhaps abusively) as ``idiom'' or ``dialect'' in English. Aside from phonetic differences, we note that there are a few hundred words that are specific to each such ``grai''. For example, lists of such words, called regionalisms, are available online\footnote{\url{http://regionalisme.ro}}. The dialects included in our dataset have lists comprising between 300 and 800 regionalisms. In summary, the Romanian dialects have several distinctive features, such as:
\begin{itemize}
    \item Phonetic differences, e.g.~the word ``ce'' (what) is pronounced ``ci'' in the Moldovenesc dialect and ``ce'' in other dialects.
    \item Regionalisms, e.g.~the word ``melon'' is translated as ``pepene'' in the Muntenesc dialect, ``harbuz'' in the Moldovenesc dialect, ``lubeni\c{t}\u{a}'' in the B\u{a}n\u{a}\c{t}ean and Oltenesc dialects, and ``curcubete'' or ``lebeni\c{t}\u{a}'' in the Ardelenesc dialect.
    \item The addition of unnecessary dialect-specific words, e.g.~``What time is it?'' is normally translated as ``C\^{a}t este ceasul?'', but in the Ardelenesc dialect, people commonly use ``Oare c\^{a}t este ceasul?'' (which can be translated as ``I wonder what time is it''). In general, it is common to use ``Oare'' when addressing a question to another person in the Ardelenesc dialect.
    \item Preference for using a different past tense in the Oltenesc and B\u{a}n\u{a}\c{t}ean dialects than in other Romanian dialects, e.g.~for the phrase ``I was'', speakers of the Oltenesc and B\u{a}n\u{a}\c{t}ean dialects dialects say ``fusei'', but the speakers of other Romanian dialects use ``am fost''.
\end{itemize}

\begin{table}[!t]
\small{
  \begin{center}
  \begin{tabular}{|l|c|}
    \hline
    {Dialect} & Population \\
    \hline
    \hline
    Ardeal (without Banat) & 5.5M\\
    Moldova & 4.2M\\
    Muntenia &	3.3M \\
    Oltenia	& 2M \\
    Banat &	1.25M \\
    Cri\c{s}ana &	1.2M \\
    Maramure\c{s} &	0.46M \\
    \hline
  \end{tabular}
  \end{center}
    \vspace{-0.4cm}
  \caption{Population size for seven of the largest regions in Romania. Our dataset includes representative dialects for the top five regions.}
\label{tab_dialect_pop}
}
\vspace{-0.1cm}
\end{table}

\noindent
\textbf{Criteria for choosing the five dialects.}
To establish the set of dialects for RoDia, we used two criteria. On the one hand, we aimed to include as many dialects as possible. On the other hand, we were limited by the low number of audio samples available for dialects spoken in small regions (by low populations). We thus selected the top five most popular dialects, which are representative for the regions depicted in Figure \ref{fig_Romania} from our paper. In Table \ref{tab_dialect_pop}, we provide the size of the population in each region of Romania corresponding to one of the top seven Romanian dialects. Dialects that are not included in RoDia correspond to smaller sub-regions, e.g.~Cri\c{s}ana and Maramure\c{s}, for which it is even harder to collect sufficient audio samples. 

\end{document}